\documentclass[runningheads]{llncs}
\usepackage[T1]{fontenc}
\usepackage{graphicx}
\usepackage{amsfonts,colortbl,xcolor}
\usepackage{amssymb}
\usepackage{multirow,amsmath,booktabs,adjustbox}
\begin{document}
\title{Fuse \& Calibrate:  A bi-directional Vision-Language Guided Framework for Referring Image Segmentation}
%
%
\author{Yichen Yan\inst{1} \and
Xingjian He\inst{1} \and
Sihan Chen\inst{2}\and Shichen Lu\inst{3}\and Jing Liu\inst{1,2} }
\authorrunning{Yichen Yan et al.}
%
\institute{Institute of Automation, Chinese Academy of Sciences, China \and
School of Artificial Intelligence, University of Chinese Academy of Sciences, China \and
School of Computer Science and Engineering, Beihang University, China
\\
\email{yanyichen2021@ia.ac.cn}\\
\email{\{xingjian.he,sihan.chen,jliu\}@nlpr.ia.ac.cn}\\
\email{sclu2020@buaa.edu.cn}}
\maketitle              
\begin{abstract}

Referring Image Segmentation (RIS) aims to segment an object described in natural language from an image, with the main challenge being a text-to-pixel correlation. Previous methods typically rely on single-modality features, such as vision or language features, to guide the multi-modal fusion process. However, this approach limits the interaction between vision and language, leading to a lack of fine-grained correlation between the language description and pixel-level details during the decoding process. In this paper, we introduce FCNet, a framework that employs a bi-directional guided fusion approach where both vision and language play guiding roles. Specifically, we use a vision-guided approach to conduct initial multi-modal fusion, obtaining multi-modal features that focus on key vision information. We then propose a language-guided calibration module to further calibrate these multi-modal features, ensuring they understand the context of the input sentence. This bi-directional vision-language guided approach produces higher-quality multi-modal features sent to the decoder, facilitating adaptive propagation of fine-grained semantic information from textual features to visual features.  Experiments on RefCOCO, RefCOCO+, and G-Ref datasets with various backbones consistently show our approach outperforming state-of-the-art methods.

\keywords{Referring Image Segmentation  \and Vision-Language Models \and Fusion \& Calibration.}
\end{abstract}
\begin{figure}[t]
  \centering
  \vspace{0pt}
  \includegraphics[width=\linewidth]{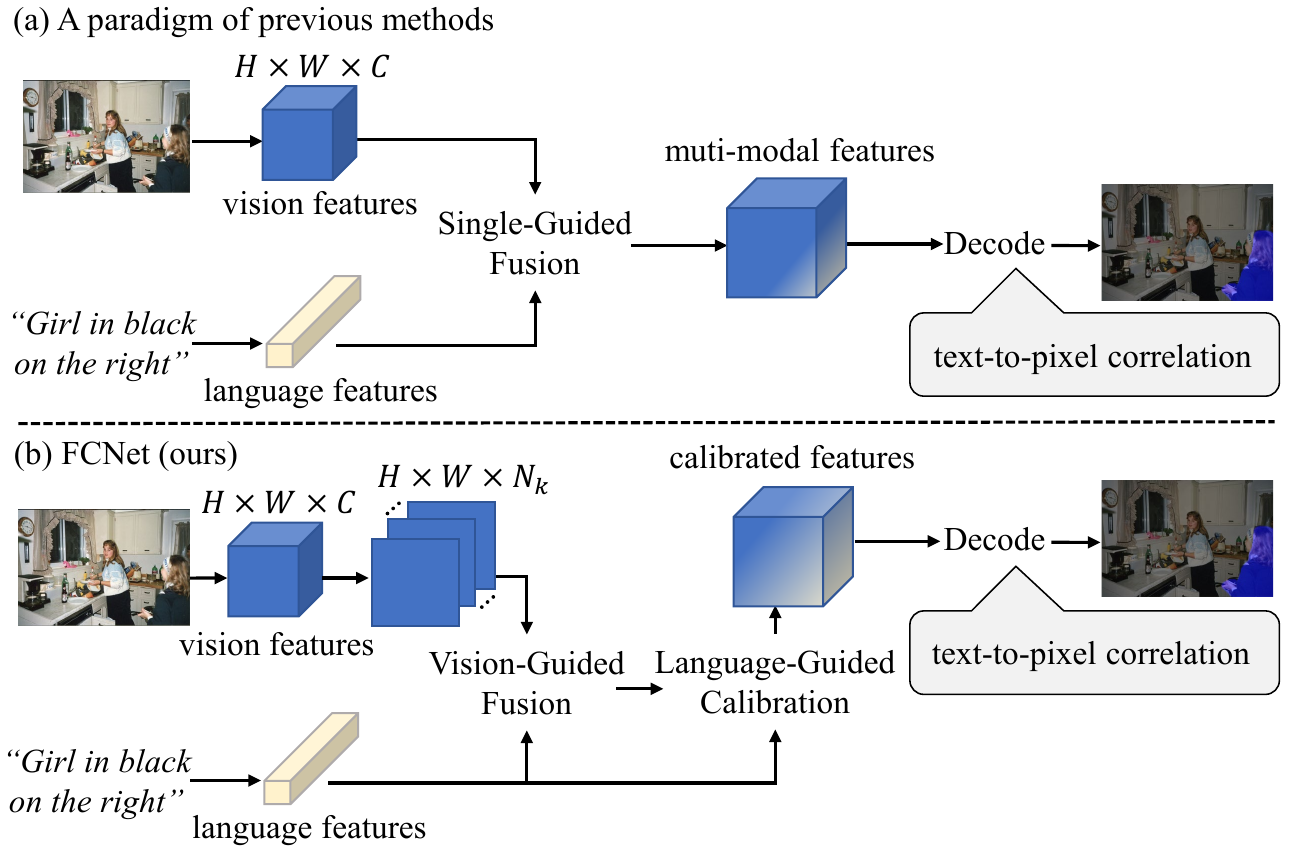}
  \caption{(a): In the previous methods (i.e., VLT \cite{ding2022vlt}), the single-guided fusion approach impede effective interaction of vision and language, resulting in a suboptimal text-pixel correlation in the decoding stage. (b): Conversely, our fusion approach includes two parts: an initial vision-guided fusion process and a language-guided calibration process. we extract the key visual information with channel $N_k$  from the original vision features and fuse them with language features by a vision-guided approach. Then we use the global language information to calibrate these fused features by a language-guided approach. This bi-directional vision-language guided approach can obtain multi-modal features where the visual and linguistic information can deeply integrate. This proves to be more advantageous for text-to-pixel correlation during the decoding stage.}
  \vspace{0pt}
  
\end{figure}

\section{Introduction}
\label{sec:intro}
Referring Image Segmentation (RIS) aims to segment an object described in natural language expressions from an image \cite{hu2016segmentation}. This task has numerous potential applications, including interactive image editing and human-object interaction \cite{wang2019reinforced}. This task is challenging due to the distinct data properties between text and image which hinders the correlation between the text and  pixel-level region.

Early works \cite{hu2016segmentation,liu2017recurrent,li2018referring} have mainly addressed this challenge by effectively fusing vision and language features, employing a common approach that leverages concatenation and convolution. With the widespread use of attention mechanisms, a series of methods \cite{ding2022vlt,kim2022restr,ye2019cross} have adopted cross-attention to learn cross-modal features effectively. However, previous methods often utilize a single-guided fusion approach, meaning they use either vision or language to guide the entire fusion process. This results in insufficient cross-modal interaction during the fusion process, leading to suboptimal text-to-pixel correlation in the decoding process.
As illustrated in Fig. 1(a), previous methods fuse vision and language features directly using a single-guided fusion approach to obtain multi-modal features. However, RIS is a multi-modal task that requires segmenting a target based on input language, making both language and vision information crucial. The single-guided fusion approach only preserves a portion of the multi-modal information. Moreover, this fusion approach lacks sufficient cross-modal interaction during the fusion process, resulting in suboptimal text-to-pixel correlation in the decoding process.

In this paper, we argue that a holistic and better understanding of multi-modal information can be achieved by fusing features guided by both modalities together. However, this is not achievable with the generic single-guided attention mechanism. Therefore, we propose a novel fusion approach that includes a initial fusion process and a calibration process guided by vision and language, respectively. As illustrated in Fig. 1(b), we propose FCNet, which first extracts the key vision features from the original image features ($H \times W \times C$). The channels of these key visual features are $N_k$ ($N_k \textless C $). We then fuse these key vision features with language features using a vision-guided approach, resulting in a series of multi-modal emphasis features that represent emphases on different channels. Next, we utilize the global language representation extracted by the text encoder to adaptively calibrate the emphasis features, ensuring that the emphasis information aligns as closely as possible with the input sentence. Through this approach, the calibrated features can comprehend the contextual information of the input sentence. Then, the calibrated features are sent to a decoder to correlate the text and pixel information effectively. FCNet leverages a bi-directional vision-language guided approach, incorporating both vision-guided fusion and language-guided calibration, thereby enhancing text-to-pixel correlation and ensuring accurate results. In summary, our contributions are: 

\begin{itemize}
    \item We propose a novel method called FCNet for the Referring Image Segmentation task, which leverages bi-directional vision-language guidance in the cross-modal fusion stage to address the text-to-pixel correlation effectively.
    \item Our proposed method introduces a novel language-guided calibration method to adaptively calibrate the multi-modal features after the fusion process, ensuring that the calibrated features understand both vision details and language context.
    \item We evaluate our method on three challenging benchmarks, and we achieve new state-of-the-art results on all datasets: RefCOCO, RefCOCO+, and G-Ref.
\end{itemize}

\section{Related Work}

\textbf{Referring Image Segmentation.} Referring Image Segmentation
is to segment a target region in an image by understanding a given language expression. This task was first introduced by \cite{hu2016segmentation}. Early works \cite{hu2016segmentation,liu2017recurrent,li2018referring} extract the vision and language features by CNN and LSTM, respectively. The extracted features are fused by concatenation to obtain the multi-modal features. Consequently, the fused vision-language features are inputted to a fully convolutional network to obtain the target segmentation mask. 

With the advent of the attention mechanism, more and more researchers proposed the attention-based framework. VLT \cite{ding2022vlt} employs a transformer-based network with encoder-decoder attention for enhancing the global context information. LAVT \cite{yang2022lavt} utilizes pre-trained BERT \cite{devlin2018bert} and Swin Transformer \cite{liu2021swin} to encode vision and language features. Recently, CRIS \cite{wang2022cris}  employs pre-trained CLIP \cite{wang2022clip} for its ability to align vision and language features.  RefSegformer \cite{wu2023towards} introduces a novel task where the described objects may not be present in the image. CrossVLT \cite{cho2023cross} leverages a stage-divided vision and language transformer encoders to perform the cross-aware early fusion and mutually enhances the robustness of each encoder.

Nevertheless, in previous methods, the fusion process is usually guided by a single modality such as image or language, leading to an insufficient interaction of the two modalities. On the contrary, we introduce a novel method named FCNet, leveraging an initial vision-guided fusion module and a language-guided calibration module to effectively fuse the cross-modal information and achieve an effective correlation of text-to-pixel in the decoding stage.

\section{METHODOLOGY}

As illustrated in Fig. 2, our proposed framework utilizes various vision and text encoder, extracting key vision features to guide the multi-modal fusion in the Emphasis Generation Module, resulting in a series of emphasis features, which represent different emphases on different vision information. Consequently, we use the global language representation to calibrate these emphasis features to make the emphasis features understand the context of the input sentence as much as possible in the Emphasis Calibration Module. After calibration, these calibrated features are fed into a transformer to decode the final result.
\begin{figure*}[t]
  \centering
  \includegraphics[width=\linewidth]{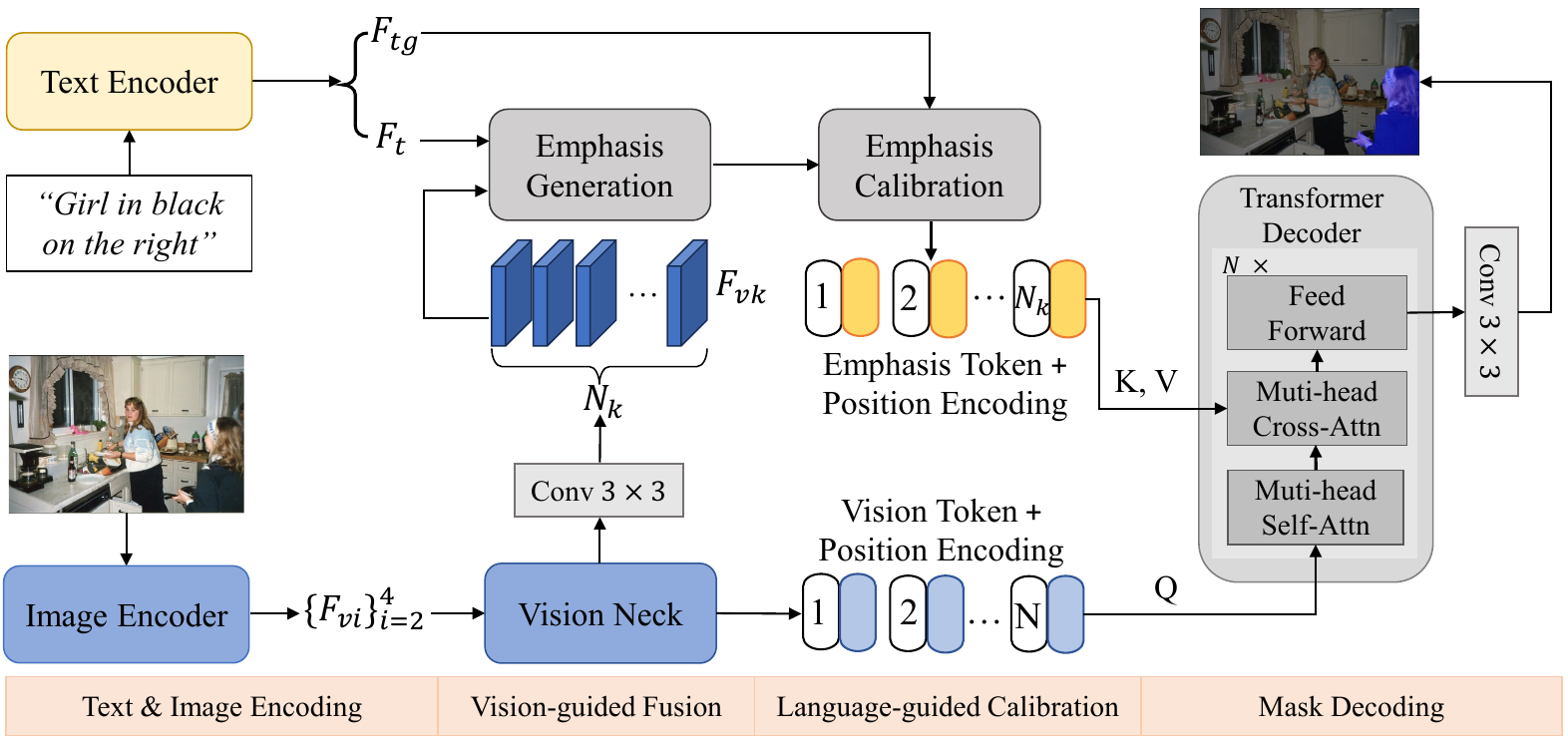}
  \caption{ There are four main stages in our method, text \& image encoding, vision-guided fusion, language-guided calibration and mask decoding. The main modules of our method are the Emphasis Generation and Emphasis Calibration.}
\end{figure*}

\subsection{Text \& Image Encoding}

\textbf{Text Encoder.} For a given sentence $T \in \mathbb{R}^{L}$, we utilize a transformer \cite{vaswani2017attention} encoder to obtain text features $F_t \in \mathbb{R}^{L \times C}$. We  use byte pair encoding (BPE) to begin the text sequence with the \texttt{[SOS]} token and end it with the \texttt{[EOS]} token. We also use the highest layer's activations of the encoder at the \texttt{[EOS]} token as the global feature for the entire language expression. The global feature represent a global context comprehension of the input sentence. Therefore, we extract this feature to help the calibration process. This feature is linearly transformed and denoted as $F_{tg} \in \mathbb{R}^{C}$. Here, $C$  represent the feature dimension, while $L$ is the length of the language expression.

\textbf{Image Encoder.} For an input image $I \in \mathbb{R}^{H \times W \times 3}$, we have two vision backbone options, ResNet \cite{he2016deep} or Swin Transformer \cite{liu2021swin}.  We utilize multiple vision features from the 2nd to 4th stages of ResNet or Swin Transformer, which are denoted as $F_{v2} \in \mathbb{R}^{\frac{H}{8} \times \frac{W}{8} \times C_2}$, $F_{v3} \in \mathbb{R}^{\frac{H}{16} \times \frac{W}{16} \times C_3}$ and $F_{v4} \in \mathbb{R}^{\frac{H}{32} \times \frac{W}{32} \times C_4}$. Here, $H$, $W$ are the input size of the image, and $C_2$, $C_3$, and $C_4$ denote the channels. 

\subsection{Vision-guided Fusion}

 In the vision-guided fusion stage, we first utilize a Vision Neck to fuse the multiple vision features $\left\{{F}_{vi}\right\}_{i=2}^{4}$ into one image feature map. An image often contains a lot of information and much of it is irrelevant to the input sentence. In our method, we utilize a convolution operation to compress these image features into a series of key vision features. We mainly focus on these key vision features and we utilize these key vision features to guide the fusion of the multi-modal features and generate a series of emphasis features, which represent different emphases on these key vision features. 

 \textbf{Vision Neck.} Vision Neck is used to fuse multiple vision features $\left\{{F}_{vi}\right\}_{i=2}^{4}$ into $F_v$. Initially we upsample $F_{v4}$ to obtain $F_{m4} \in \mathbb{R}^{\frac{H}{16} \times \frac{W}{16} \times C}$ by the following equation: 
 \begin{equation}
F_{m 4}=Up(\sigma(F_{v4}W_{v4})),
 \end{equation}
 
 Here, $Up(\cdot)$ denotes $2\times$ upsampling function, and $\sigma(\cdot)$ denotes ReLU activation function. Subsequently, we obtain the vision features $F_{m3}$  and $F_{m2}$ by the following equations:
\begin{equation}
\begin{aligned}
& F_{m_3}=\left[\sigma\left(F_{m_4} W_{m_4}\right), \sigma\left(F_{v_3} W_{v_3}\right)\right], \\
& F_{m_2}=\left[\sigma\left(F_{m_3} W_{m_3}\right), \sigma\left(F_{v_2}^{\prime} W_{v_2}\right)\right], \\
& F_{v_2}^{\prime}=Avg\left(F_{v_2}\right),
\end{aligned}
\end{equation}
Where $Avg(\cdot)$ denotes a kernel size of 2 $\times$ 2 average pooling operation, $[,]$ denotes the concatenation.Subsequently, we concatenate the three multi-modal features ($F_{m 4}, F_{m 3}, F_{m 2}$) and use a $1 \times 1$ convolution layer to aggregate them to obtain vision features $F_{m} \in \mathbb{R}^{\frac{H}{16} \times \frac{W}{16} \times C}$. Then, we obtain the 2D spatial coordinate feature $F_{coord} \in \mathbb{R}^{\frac{H}{16} \times \frac{W}{16} \times C}$ and concatenate it with $F_{m}$. The details are as follows: 
\begin{equation}
\begin{aligned}
& F_m = Conv\left(\left[F_{m_2}, F_{m_3}, F_{m_4}\right]\right), \\
& F_v = Conv\left(\left[F_m,F_{coord}\right]\right),
\end{aligned}
\end{equation}

Here, $F_v\in \mathbb{R}^{H_{v} \times W_{v} \times C}$, $H_v=\frac{H}{16}$, $W_v = \frac{W}{16}$, is the vision features used in the following process.

\textbf{Emphasis Generation.} In the original vision features $F_v$, there exists redundant visual information that impedes the mutual understanding between visual and linguistic components. Notably, vision features $F_v$ often comprise a substantial number of channels $C$. In FCNet, we employ a $3 \times 3$ convolution to extract key vision features, reducing the dimension from $C$ to $N_k$, resulting in $F_{vk} \in \mathbb{R}^{H_{v} \times W_{v} \times N_k}$. Here, $N_k$ serves as a hyperparameter representing the number of key vision features on which we intend to focus. We send these key vision features into the Emphasis Generation module to guide the initial fusion of multi-modal features. We flatten these key vision features and use the cross-attention mechanism with language features $F_t$ to generate a series of emphasis features.
\begin{equation}
\begin{aligned}
& F_{vk}^{\prime}=Flatten(F_{vk}), \\
& F_e = CrossAttn(F_{vk}^{\prime},F_t,F_t),
\end{aligned}
\end{equation}
Here, $F_{vk}^{\prime }\in \mathbb{R}^{\left(H_{v}W_{v}\right) \times N_k}$. In the cross-attention mechanism, Q is derived from $F_{vk}^{\prime}$ to query the language features, while K and V are extracted from $F_t$ to respond to key visual information. The dimension of the attention map is $N_k \times L$, indicating the importance of each word in the input sentence with respect to each key vision feature. We obtain the emphasis features $F_e \in \mathbb{R}^{N_k \times C}$. Each vector in $F_e$, with a dimension of $1 \times C$, signifies a distinct comprehension and emphasis of the vision features. These emphasis features are multi-modal representations containing key vision information and comprehensive language details.

\begin{figure}[thbp]
  \centering
  \vspace{0pt}
  \includegraphics[width=0.85\linewidth]{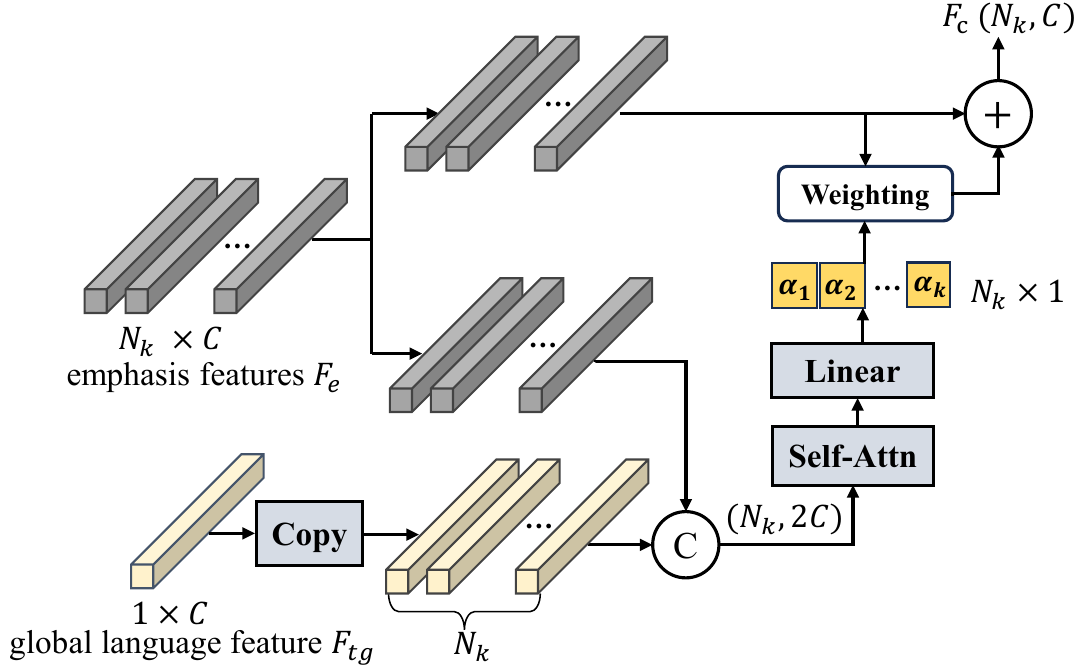}
  \caption{The process of our language-guided calibration. We employ the global language representation $f_{vg}$ to guide the generation of corresponding scores for each emphasis feature.}
  \vspace{0pt} 
\end{figure}

\subsection{Language-guided Calibration}

In previous approaches, fused multi-modal features were directly input into the decoder for the final mask decoding. However, these multi-modal features obtained by the single-guided approach lack sufficient interaction between vision and language, hindering the propagation of fine-grained semantic information from textual features to visual features during the decoder stage. In FCNet, we utilize global language representation to guide the calibration process in the Emphasis Calibration module.

As shown in Fig. 3, we create $N_k$ copies of $F_{tg}$ and concatenate them with emphasis features $F_e$. Subsequently, we apply a self-attention operation to the concatenated features, facilitating the widespread dissemination of global language information among the emphasis features. Then, we employ a linear transformation to produce a series of scalars $\left\{{\alpha}_{i}\right\}_{i=1}^{N_k}$. For $i$-th emphasis $f_{ei} \in F_e$, ($i = 1, 2...N_k$), $\alpha_i$ denotes relation to the global language information and its significance. We employ the set of scalars $\left\{{\alpha}_{i}\right\}_{i=1}^{N_k}$ to balance the emphasis features through a weighting operation. Subsequently, we utilize these weighted features to calibrate the input emphasis features by addition, yielding a series of calibrated emphasis features $F_c \in \mathbb{R}^{N_k \times C}$, which are also called calibrated features. The calibrated features focus on key visual information while fully understanding the context of the language.

\subsection{Mask Decoding}

We leverage a standard transformer \cite{vaswani2017attention} decoder to adaptively propagate
fine-grained text-level information from textual features. As illustrated in Fig. 2, the transformer takes calibrated emphasis features $F_c$ and vision features $F_v$ as input. In our decoder, each layer consists of a multi-head self-attention layer, a multi-head cross-attention layer, and a feed-forward network. The specific process are as follows: 
\begin{equation}
\begin{aligned}
& F^{\prime}_v=MHSA(LN(F_v))+F_v,\\
& F^{\prime}_q=MHCA(LN(F^{\prime}_v),F_c)+F^{\prime}_v,\\
& F_q=MLP(LN(F^{\prime}_q))+F^{\prime}_q,
\end{aligned}
\end{equation}
Here, $F^{\prime}_v$ is the intermediate  features, $MHSA(\cdot)$ and $LN(\cdot)$ denote the multi-head self-attention layer and Layer Normalization, respectively. $MHSA(\cdot)$ and $MLP(\cdot)$ denote the multi-head cross-attention layer and the MLP block. The final mask is obtained by a $3\times3$ convolution layer. The model is optimized with binary cross-entropy loss.

\begin{table*}[thbp]
    \setlength{\belowcaptionskip}{1.0pt}
    \begin{center}
    \caption{\textbf{Comparisons with the state-of-the-art approaches on three benchmarks. IoU is utilized as the metric.}
    }
    \begin{adjustbox}{width=0.95\textwidth}
    \begin{tabular}{l|c|ccc|ccc|cc}
        \toprule[1.2pt]
        \multirow{2}{*}{Method} & \multirow{2}{*}{Vision Backbone}   & \multicolumn{3}{c|}{RefCOCO} & \multicolumn{3}{c|}{RefCOCO+} & \multicolumn{2}{c}{G-Ref} \\
        \cline{3-10}
        ~ & ~  &val & test A & test B & val & test A & test B & val & test \\
        \hline

        MAttNet \cite{yu2018mattnet}           & ResNet-101 &  56.51 & 62.37 & 51.70 & 46.67 & 52.39 & 40.08 & 47.64 & 48.61 \\
        EFNet \cite{feng2021encoder}           & ResNet-101  &62.76 & 65.69 & 59.67 & 51.50 & 55.24 & 43.01 & - & - \\
        CGAN \cite{luo2020cascade}                          &ResNet-101 &64.86& 68.04& 62.07& 51.03& 55.51& 44.06& 51.01& 51.69\\
        SeqTR \cite{li2021referring}                 & ResNet-101   &67.26& 69.79& 64.12& 54.14& 58.93& 48.19& 55.67& 55.64 \\
                RefTr \cite{li2021referring}                & ResNet-101   &\underline{70.56} &\underline{73.49}& \underline{66.57}& 61.08& 64.69& 52.73& 58.73& 58.51 \\
        CRIS \cite{wang2022cris}                        & ResNet-101  & 70.47 & 73.18 & 66.10 & \underline{62.27} &\underline{ 68.08} & \underline{53.68} & \underline{59.87} & \underline{60.36} \\

         \rowcolor{lightgray}FCNet(Ours)                     &ResNet-101 & \textbf{71.29} & \textbf{74.49} & \textbf{66.70} & \textbf{63.35} & \textbf{68.41} & \textbf{54.79} & \textbf{60.88} &\textbf{ 61.39} \\
        \midrule

        LAVT \cite{yang2022lavt}                           & Swin-Base  & 72.73 & 75.82 & 68.79 & 62.14 & 68.38 & 55.10 & 61.24 & 62.09 \\
         VLT+ \cite{ding2022vlt}                & Swin-Base& 72.96& 75.56& 69.60 &63.53& 68.43& 56.92& \underline{63.49}& \underline{66.22}\\
        
         RefSegformer \cite{wu2023towards}    & Swin-Base& 73.22& 75.64& 70.09 &63.50& 68.69& 55.44& 62.56& 63.07\\
         CrossVLT \cite{cho2023cross}   & Swin-Base&73.44& 76.16& \underline{70.15}& \underline{63.60}& \underline{69.10}& 55.23& 62.68& 63.75\\
        PVD \cite{cheng2023parallel} & Swin-Base&\underline{74.82}& \underline{77.11}& 69.52& 63.38 &68.60& \underline{56.92}& 63.13& 63.62\\
    
        \rowcolor{lightgray}FCNet(Ours)                     &Swin-Base &\textbf{75.02} & \textbf{77.49} & \textbf{71.47} & \textbf{66.38} & \textbf{71.25} & \textbf{58.71} & \textbf{65.22} & \textbf{66.49}\\
        \bottomrule[1.2pt]
    \end{tabular}
    \end{adjustbox}
    \label{tab:sota}
    \end{center}
    \vspace{-5.0mm}
\end{table*}

\section{Experiments}

\subsection{Implementation Details}

\textbf{Datasets.} We conduct our experiments on three datasets, RefCOCO \cite{kazemzadeh2014referitgame}, RefCOCO+ \cite{kazemzadeh2014referitgame} and G-Ref \cite{nagaraja2016modeling}. The RefCOCO dataset comprises 142,209 language expressions describing 50,000 objects within 19,992 images, while the RefCOCO+ dataset includes 141,564 language expressions corresponding to 49,856 objects in 19,992 images. The primary distinction between RefCOCO and RefCOCO+ is that RefCOCO+ omits words indicating location properties (such as left, top, front) in expressions. G-Ref has 104,560 referring language expressions associated with 54,822 objects of 26,711 images. The language usage in the G-Ref is more casual and complex, and the sentence length of G-Ref is also longer.

\textbf{Metrics.} Following previous works \cite{ding2022vlt,wang2022cris}, we adopt two metrics to verify the effectiveness: IoU and Precision@X. The IoU calculates intersection regions over union regions of the predicted segmentation mask and the ground truth. The Precision@X measures the percentage of test images with an IoU score higher than the threshold $\mathbf{X} \in \left\{0.5, 0.6, 0.7, 0.8, 0.9\right\}$, which focuses on the location ability of the method.

\textbf{Experiment Settings.} Following the previous work \cite{wang2022cris,yang2022lavt}, we use the same experiment settings. Our framework utilizes ResNet-101 \cite{he2016deep} and Swin-Base \cite{liu2021swin}. The image size is set to $480 \times 480$ input sentences are set with a maximum sentence length of 17 for RefCOCO and RefCOCO+, and 22 for G-Ref. Each Transformer Decoder layer has 8 heads, and the feed-forward hidden dimension is set to 2048. We train the network for 50 epochs using the Adam optimizer with the learning rate lr = 0.0001. The learning rate is decreased by a factor of 0.1 at the 35th epoch. We train the model with a batch size of 64 on 8 RTX Titan with 24 GPU VRAM.

\subsection{Comparison with State-of-the-art Methods }
In Table 1, we evaluate FCNet against the other methods on the three datasets using the different vision backbone. We first use the basic visual backbone ResNet-101. Results show that our method outperforms other methods using the same backbone, including the CRIS \cite{wang2022cris} and RefRr \cite{li2021referring}. Then we utilize Swin-Base as backbone, and results show that our method outperforms other methods on all datasets.
Compared to the second-best performing method PVD \cite{cheng2023parallel}, FCNet achieves absolute margins of 1.20, 0.38, and 1.95 scores on the validation, testA, and testB subsets of RefCOCO, respectively. Our proposed method also outperforms $\rm{M^3}$AII on the  RefCOCO+ dataset with 3.00, 2.65 and 1.79 absolute score improvements. On the G-Ref dataset, our method surpasses the second-best methods on the validation and test subsets from the UMD partition by absolute margins of 2.09 and 2.87 on the validation and test subsets.

\section{Ablation Study}
We conduct several ablations to evaluate the effectiveness of the
key components in our proposed network, including the vision-guided fusion stage and language-guided calibration stage. We do the ablation study on the testA split of RefCOCO+, we use ResNet-50 as the vision backbone. We first conduct a comprehensive comparison of our proposed Emphasis Generation Module and Emphasis Calibration Module which comprehensively evaluate their effectiveness. Detailed ablations of each part are provided in the following paragraphs.

\begin{table}[htbp]
    \begin{center}
    \caption{\textbf{Comprehensive comparison of Emphasis Generation Module (EGM) and Emphasis Calibration Module (ECM).}}
    \setlength{\tabcolsep}{1mm}{
    \begin{tabular}{c|c|c|c|c|c|c|c}
        \toprule[1.2pt]
        EGM&ECM & IoU & Pr@50 & Pr@60 & Pr@70&Pr@80&Pr@90 \\
        \midrule
        \checkmark & \checkmark & \textbf{68.04} & \textbf{80.07} & \textbf{76.49} & \textbf{69.22} & \textbf{53.04} & \textbf{15.13} \\
        \checkmark& ~ &  66.79 & 78.79 & 74.64 & 68.41 & 52.18 & 15.04 \\

            ~ & ~ & 66.03 & 77.48 & 73.87 & 67.96 & 51.83 & 14.75\\
        \bottomrule[1.2pt]
    \end{tabular}
    \label{tab:Nq}}
    \end{center}
\end{table}

\textbf{Comprehensive Comparison}. We conduct the comprehensive comparison to directly demonstrate the effectiveness of our proposed vision-guided Emphasis Generation Module (EGM) and language-guided Emphasis Calibration Module (ECM). As illustrated in Table 2, we remove EGM and ECM in turn, removing the ECM results in a decrease of 1.25; removing both components yields an even more inferior result, with a decrease of 2.01. This comprehensive comparison validates the significance of bi-directional vision-language guidance, and adopting both together can synergistically enhance results.

\begin{table}[htbp]
    \setlength{\belowcaptionskip}{0pt}
    \begin{center}
    \caption{\textbf{Influence of $N_k$.}}
    \setlength{\tabcolsep}{1.8mm}{
    \begin{tabular}{c|c|c|c|c|c|c}
        \toprule[1.1pt]
        $N_k$ & IoU & Pr@50 & Pr@60 & Pr@70 & Pr@80 & Pr@90 \\
        \hline
          32 & 67.25 & 79.65 & 75.48 & 68.51 & 52.60 & 14.94 \\
          24 & 67.56 & 79.96 & 76.83 & 69.03 & 52.91 & 14.88 \\
          16 & \textbf{68.04} & \textbf{80.07} & \textbf{76.49} & \textbf{69.22}  & \textbf{53.04}  & \textbf{15.13} \\
          8  & 67.46 & 78.89 & 75.61 & 68.15 & 52.71 & 14.53 \\
          4  & 66.23 & 78.02 & 74.69 & 68.48 & 52.23 & 14.29 \\
          2  & 65.75 & 77.50 & 74.16 & 68.21 & 51.88 & 14.31 \\
          1  & 65.36 & 77.61 & 72.67 & 65.89 & 51.61 & 13.88 \\
        \bottomrule[1.1pt]
    \end{tabular}
    \label{tab:Nq}}
    \end{center}
    \vspace{0.0mm}
\end{table}

\textbf{Channel of Key Vision Features.} $N_k$ is the channels of the key vision features which also decide how many emphasis features we generate. In order to clarify the influence of the $N_k$, we set the $N_k$ to a series of different numbers. The
results are reported in Table 3.  Based on the results, it is evident that choosing an excessively large or small value for $N_k$ yields suboptimal results, highlighting the importance of an appropriate balance in extracting key visual information. Consequently, we have set $N_k$ to 16 for its superior performance.

\begin{table}[htbp]
    \centering
    \caption{\textbf{Ablation study on language-guided calibration, ECM denotes the Emphasis Calibration module.}}
    \setlength{\tabcolsep}{1mm}
    \begin{tabular}{c|c|c|c|c|c|c|c}
        \toprule[1.1pt]
        $N_k$ & ECM & IoU & Pr@50 & Pr@60 & Pr@70 & Pr@80 & Pr@90 \\
        \hline
        \multirow{2}{*}{24} & \checkmark & 67.56 & 79.96 & 76.83 & 69.03 & 52.91 & 14.89 \\
        \cline{2-8}
        ~ & -   & 66.42 & 78.34 & 74.78 & 67.85 & 50.70 & 13.79 \\
        \hline
        \multirow{2}{*}{16} & \checkmark & \textbf{68.04} & \textbf{80.07} & \textbf{76.49} & \textbf{69.22} & \textbf{53.04} & \textbf{15.13} \\
        \cline{2-8}
        ~ & -  & 66.79 & 78.79 & 74.64 & 68.41 & 52.18 & 15.04 \\
        \hline
        \multirow{2}{*}{8} & \checkmark & 67.46 & 77.50 & 74.16 & 68.21 & 51.88 & 14.31 \\
        \cline{2-8}
        ~ & -   & 66.19 & 77.41 & 73.89 & 67.27 & 50.93 & 13.45 \\
        \bottomrule[1.1pt]
    \end{tabular}
    \label{tab:Nq}
\end{table}

\textbf{Language-guided Calibration.} The core of our method is using the language-guided calibration approach to adaptively calibrate the emphasis features. To demonstrate its effectiveness, we conducted a comparative experiment across various settings of $N_k$. In this experiment, we omitted the Emphasis Calibration module, allowing the emphasis features to be directly fed into the decoder. As shown in Table 4, the results indicate that, across different $N_k$ settings, the Emphasis Calibration module consistently achieves superior results, demonstrating the effectiveness of our proposed language-guided approach.

\begin{figure}[ht]
  \centering
  \vspace{0pt}
  \includegraphics[width=\linewidth]{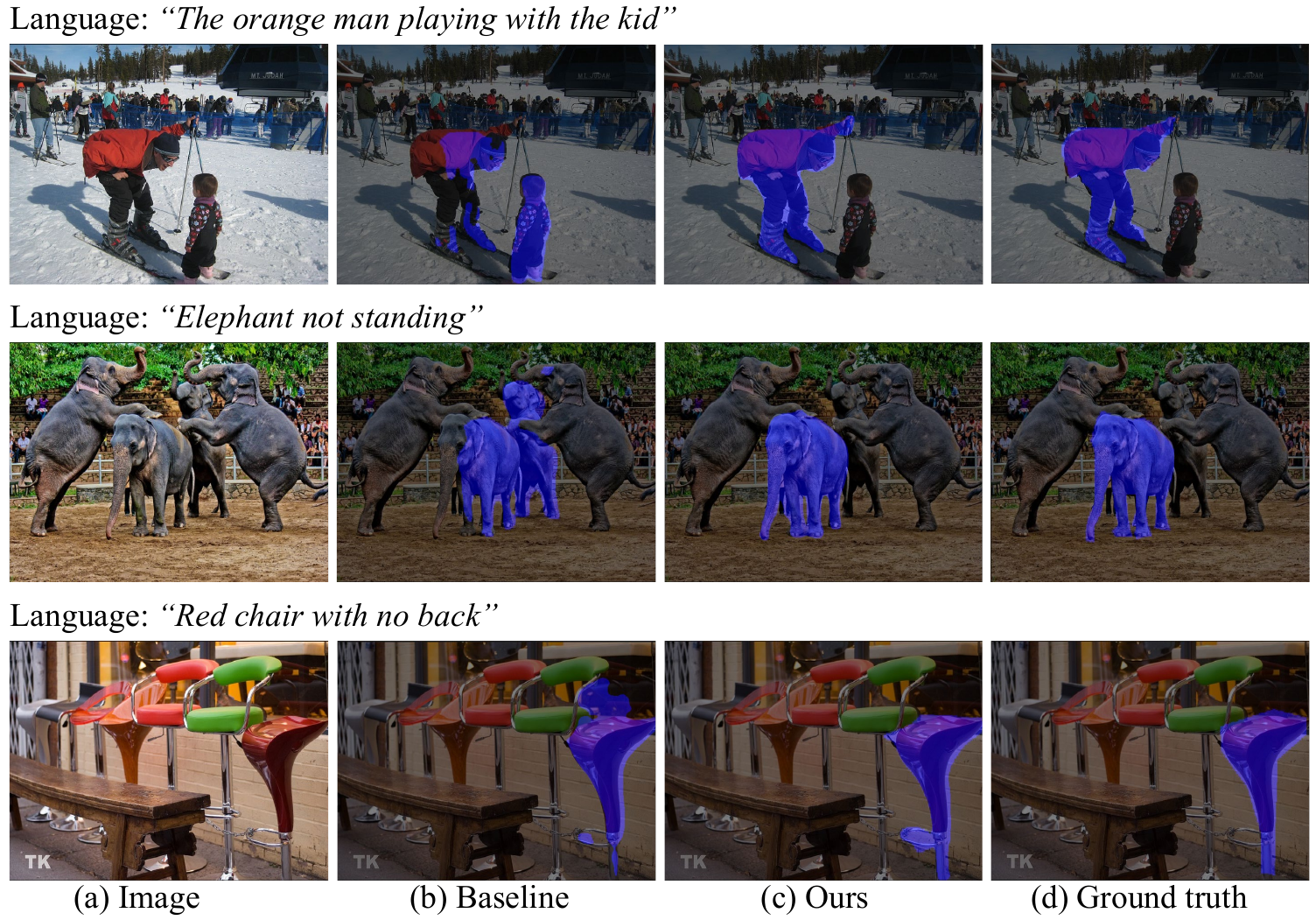}
  \caption{Visualization comparison of FCNet and  baseline. The baseline framework utilize the single-guided fusion approach.}
  \vspace{0pt}
\end{figure}

\section{Visualization}
In this section, we present visualization results comparing FCNet with our baseline, which utilizes the single-guided fusion approach. We show the mask predictions of three different input language expressions for every example. As illustrated in Fig. 4, the first and second rows show the capability of our method to identify specific targets by correlating key pixel regions with attribute words, such as "orange man" and "not standing."  The third row demonstrates FCNet's superior performance in delivering accurate segmentation results compared to our baseline. This is because our method obtains high-quality multi-modal features by the bi-directional vision-language guided fusion and calibration approach. This ensures that the both visual information and linguistic information can effectively interact with each other , leading to a more effective and accurate text-to-pixel correlation.

\section{Efficiency Comparison}

 In Table 5, we provide the parameters and GFLOPs of different models, such as LAVT \cite{yang2022lavt} and CRIS \cite{wang2022cris}. We also include training times using 8 RTX Titan GPUs, along with inference times. Both CRIS and our method utilize ResNet101 \cite{he2016deep} as the vision backbone, while LAVT employs Swin-Base \cite{liu2021swin} which is the stronger backbone than ours. The results are all from the RefCOCO+ testA split. The obtained results indicate that our method not only achieves superior performance but also exhibits relatively faster training and inference speeds in comparison by only using a less power vision backbone.

\begin{table}[htbp]
     \setlength{\abovecaptionskip}{0pt} 
    \setlength{\belowcaptionskip}{0pt} 
    \begin{center}
    \vspace{-3pt}
    \caption{\textbf{Efficiency comparison with other open-source models, “FPS” denotes the inference speed.}}
    \setlength{\tabcolsep}{1.8mm}{
    \begin{tabular}{p{0.80cm}|c|c|c|c|c}
        \toprule[1.2pt]
        Model & IoU & GFLOPs & time & Params & FPS  \\
        \midrule
          CRIS & 62.27 & 79.45 & 18.2h & 188.85M  &   12.07 \\
          LAVT & 62.14 & 259.61  & 46.5h & 203.70M  & 5.23   \\
          Ours & 63.35 & 84.05 & 18.6h & 190.04M & 12.14  \\
        \bottomrule[1.2pt]
        
    \end{tabular}
    \label{tab:Nq}}
        \vspace{0pt}
    \end{center}
 
\end{table}

\section{Conclusion} 
In this paper, we introduce a novel framework for RIS task named FCNet, which effectively fuses and calibrates the multi-modal features by a bi-directional vision-language guided approach. Compared with previous single-guide methods, our framework achieves superior vision-language interaction in the cross-modal fusion process. Specifically, we extract key vision features from the original vision features and use them to guide the generation of a series of emphasis features that represent different emphases of the vision information. Then we utilize the global language representation to guide the calibration of these emphasis features. These calibrated features are sent to the decoder to achieve text-to-pixel correlation. Our experiments show that FCNet significantly outperforms previous state-of-the-art methods on RefCOCO, RefCOCO+ and G-Ref datasets without any post-processing. 

\section{Acknowledgments}
This work was supported by the National Science and Technology Major Project (No.2022ZD0118801), National Natural Science Foundation of China (U21B2043, 62206279).
%
%
%
\bibliographystyle{splncs04}
\bibliography{mybibliography}
%




\end{document}